\documentclass{article}

\usepackage{arxiv}

\usepackage[utf8]{inputenc} 
\usepackage[T1]{fontenc}    
\usepackage{hyperref}       
\usepackage{url}            
\usepackage{booktabs}       
\usepackage{amsfonts}       
\usepackage{nicefrac}       
\usepackage{microtype}      
\usepackage{lipsum}		
\usepackage{graphicx}
\usepackage{doi}
\usepackage{algorithm}
\usepackage{algorithmic}
\usepackage{amsmath,amssymb,amsfonts}
\usepackage{graphicx}
\usepackage{textcomp}
\usepackage{caption} 
\usepackage{subcaption}
\usepackage{mathtools}
\usepackage{float}
\usepackage{booktabs,siunitx}
\usepackage{makecell}
\usepackage{multirow}
\usepackage{xcolor}

\title{SAU: Smooth activation function using convolution with approximate identities}

\author{ 
Koushik Biswas\\
	\And
	Sandeep Kumar\\
	\And
	 Shilpak Banerjee\\
	\And
	 Ashish Kumar Pandey\\
}

\date{}

\renewcommand{\shorttitle}{}

\hypersetup{
pdftitle={SMOOTH ACTIVATION FUNCTION},
pdfsubject={Activation Function},
pdfauthor={Koushik Biswas},
pdfkeywords={Activation Function},
}

\begin{document}
\maketitle

\begin{abstract}
Well-known activation functions like ReLU or Leaky ReLU are non-differentiable at the origin. Over the years, many smooth approximations of ReLU have been proposed using various smoothing techniques. We propose new smooth approximations of a non-differentiable activation function by convolving it with approximate identities. In particular, we present smooth approximations of Leaky ReLU and show that they outperform several well-known activation functions in various datasets and models. We call this function Smooth Activation Unit (SAU). Replacing ReLU by SAU, we get 5.12\% improvement with ShuffleNet V2 (2.0x) model on CIFAR100 dataset.
\end{abstract}
\section{Introduction}
Deep networks form a crucial component of modern machine learning. Non-linearity is introduced in such networks by the use of activation functions, and the choice has a substantial impact on network performance and training dynamics. Designing a new novel activation function is a difficult task. Handcrafted activations like Rectified Linear Unit (ReLU) \cite{relu}, Leaky ReLU \cite{lrelu} or its variants are a very common choice for activation functions and exhibits promising performance on different deep learning tasks. There are many activations that have been proposed so far and some of them are ELU \cite{elu}, Parametric ReLU (PReLU) \cite{prelu}, Swish \cite{swish}, Tanhsoft \cite{tanhsoft}, EIS \cite{eis}, PAU \cite{pau}, OPAU \cite{opau}, ACON \cite{acon}, ErfAct \cite{erf}, Mish \cite{mish}, GELU \cite{gelu} etc. Nevertheless, ReLU remains the favourite choice among the deep learning community due to its simplicity and better performance when compared to Tanh or Sigmoid, though it has a drawback known as dying  ReLU in which the network starts to lose the gradient direction due to the negative inputs and produces zero outcome. In 2017, Swish \cite{swish} was proposed by the Google brain team. Swish was found by automatic search technique, and it has shown some promising performance across different deep learning tasks.

Activation functions are usually handcrafted. PReLU \cite{prelu} tries to overcome this problem by introducing a learnable negative component to ReLU \cite{relu}. Maxout \cite{maxout}, and Mixout \cite{mixout} are constructed with piecewise linear components, and theoretically, they are universal function approximators, though they increase the number of parameters in the network. Recently, meta-ACON \cite{acon}, a smooth activation has been proposed, which is the generalization of the ReLU and Maxout activations and can smoothly approximate Swish. Meta-ACON has shown some good improvement on both small models and highly optimized large models. PAU \cite{pau}, and OPAU \cite{opau} are two promising candidates for trainable activations, which have been introduced recently based on rational function approximation.

In this paper, we introduce a smooth approximation of known non-smooth activation functions like ReLU or Leaky ReLU based on the approximation of identity. Our experiments show that the proposed activations improve the performance of different network architectures as compared to ReLU on different deep learning problems.

\section{Mathematical formalism} 
\subsection{Convolution} Convolution is a binary operation, which takes two functions $f$ and $g$ as input, and outputs a new function denoted by $f*g$. Mathematically, we define this operation as follows
\begin{align}\label{e:conv}
(f*g)(x) = \int_{-\infty}^{\infty} f(y)g(x-y) \,dy.
\end{align}
The convolution operation has several properties. Below, we will list two of them which will be used larter in this article.
\begin{enumerate}
    \item[P1.] $(f\ast g)(x)=(g\ast f)(x)$,
    \item[P2.] If $f$ is $n$-times differentiable with compact support over $\mathbb{R}$ and $g$ is locally integrable over $\mathbb{R}$ then $f\ast g$ is at least $n$-times differentiable over $\mathbb{R}$. 
\end{enumerate}
Property P1 is an easy consequence of definition \eqref{e:conv}. Property P2 can be easily obtained by moving the derivative operator inside the integral. Note that this exchange of derivative and integral requires $f$ to be of compact support. An immediate consequence of property P2 is that if one of the functions $f$ or $g$ is smooth with compact support, then $f\ast g$ is also smooth. This observation will be used later in the article to obtain smooth approximations of non-differentiable activation functions.



\subsection{Mollifier and Approximate identities} 
A smooth function $\phi$ over $\mathbb{R}$ is called a mollifier if it satisfies the following three properties:
\begin{enumerate}
     \item It is compactly supported.
     \item $\int_{\mathbb{R}} \phi(x) \,dx = 1.$
    \item $ {\displaystyle  \lim _{\epsilon \to 0} \phi _{\epsilon }(x):=\lim _{\epsilon \to 0}\frac{1}{\epsilon}\phi (x/\epsilon )=\delta (x)}$, where ${\displaystyle \delta (x)}$ is the Dirac delta function.
\end{enumerate}

We say that a mollifier $\phi$ is an approximate identity if for any locally integrable function $f$ over $\mathbb{R}$, we have
\begin{align*}
{\displaystyle \lim _{\epsilon \to 0}(f\ast \phi _{\epsilon })(x)=f(x) \; \text{pointwise for all }x. }
\end{align*}


\subsection{Smooth approximations of non-differentiable functions}
Let $\phi$ be an approximate identity. Choosing $\epsilon=1/n$ for $n\in \mathbb{N}$, one can define
\begin{align}\label{e:phin}
\phi_n(x):=n\phi(n x).
\end{align}
Using the property of approximate identity, for any locally integrable function $f$ over $\mathbb{R}$, we have
\begin{align*}
{\displaystyle \lim _{n \to \infty}(f\ast \phi_n)(x)=f(x) \; \text{pointwise for all }x. }
\end{align*}
That is, for large enough $n$, $f\ast \phi_n$ is a good approximation of $f$. Moreover, since $\phi$ is smooth, $\phi_n$ is smooth for each $n\in \mathbb{N}$ and therefore, using property P2, $f\ast \phi_n$ is a smooth approximation of $f$ for large enough $n$.

Let $\sigma:\mathbb{R} \to \mathbb{R}$ be any activation function. Then, by definition, $\sigma$ is a continuous and hence, a locally integrable function. For a given approximate identity $\phi$ and $n\in \mathbb{N}$, we define a smooth approximation of $\sigma$ as $\sigma\ast \phi_n$, where $\phi_n$ is defined in \eqref{e:phin}.

\section{Smooth Approximation Unit (SAU)}
Consider the Gaussian function
\begin{align*}
    \phi(x) = \frac{1}{\sqrt{2\pi}}e^{-\frac{x^2}{2}}
\end{align*}
which is a well known approximate identity. Consider the Leaky Rectified Linear Unit (Leaky ReLU) activation function
\begin{align*}
    \operatorname{Leaky ReLU}[\alpha](x)=\begin{cases}
   x & x\geq 0\\
    \alpha x & x<0
    \end{cases}
\end{align*}
Note that $\operatorname{Leaky ReLU}[\alpha]$ activation function is hyper-parametrized by $\alpha$ and it is non-differentiable at the origin for all values of $\alpha$ except $\alpha=1$. For $\alpha=0$ and $\alpha=0.01$, $\operatorname{Leaky ReLU}[\alpha]$ reduces to well known activation functions ReLU and Leaky ReLU, respectively. For a given $n\in \mathbb{N}$, and $\alpha\neq 1$, a smooth approximation of $\operatorname{Leaky ReLU}[\alpha]$ is given by
\begin{align}\label{eq1}
   \text{SAU} = G(x,\alpha,n)= (\operatorname{Leaky ReLU}[\alpha]*\phi_n)(x)=\frac{1}{2n}\sqrt{\frac{2}{\pi}}e^{\frac{-n^2x^2}{2}}+\frac{(1+\alpha)}{2}x\notag \\
   +\frac{(1-\alpha)}{2}x \ \operatorname{erf}\left(\frac{nx}{\sqrt{2}}\right).
  \end{align}
where $\operatorname{erf}$ is the Gaussian error function
\begin{align*}
    \operatorname{erf}(x) = \dfrac{2}{\sqrt{\pi}} \int_{0}^{x} e^{-t^2} \,dt. 
\end{align*}
It is noticeable that this function is not zero centred but passes by an extremely close neighbourhood of zero. To make the function zero centered, we have multiplied the first term of (3) by a linear component $x$. To further investigate these two functions as a possible candidates for activation function, we have conducted several experiments on MNIST \cite{mnist}, CIFAR10 \cite{cifar10}, and CIFAR100 \cite{cifar10} datasets with PreActResNet-18 \cite{preactresnet}, VGG-16 (with batch-normalization) \cite{batch} \cite{vgg}, and DenseNet-121 \cite{densenet} models, and we notice that  both of them performs almost similar in every cases. So, for the rest of the paper, we will only consider the approximate identity of Leaky ReLU ($\alpha=0.15$) given in (3) as the activation function. We call this function Smooth Approximation Unit (SAU). 

We note that in GELU \cite{gelu} paper, a similar idea is utilized to obtain their activation functions. They use the product of $x$ with the cumulative distribution function of a suitable probability distribution (see \cite{gelu} for further details).


\begin{figure}[!t]
   
        \centering
    
        \includegraphics[width=12.5cm,height=6.95cm,keepaspectratio]{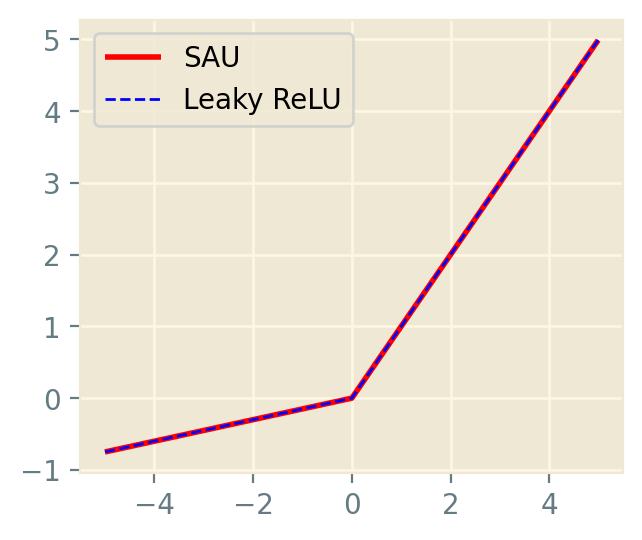}
        \vspace{0.2cm}
        \caption{Approximation of Leaky ReLU ($\alpha = 0.15$) using SAU.}
        \label{sau1}
   
\end{figure}

\subsection{Learning activation parameters via back-propagation} Back-propagation algorithm \cite{backp} and gradient descent is used in neural networks to update Weights and biases. Parameters in trainable activation functions are updated using the same technique. The forward pass is implemented in both Pytorch \cite{pytorch} \& Tensorflow-Keras \cite{keras} API, and automatic differentiation will update the parameters. Alternatively, CUDA \cite{cuda} based implementation (see \cite{lrelu}) can be used and the gradients of equation (3) for the input $x$ and the parameter $\alpha$ can be computed as follows:
\begin{align}
    \frac{\partial G}{\partial x} = \frac{-n^2x}{2n}\sqrt{\frac{2}{\pi}}e^{\frac{-n^2x^2}{2}}+\frac{(1+\alpha)}{2}
   +\frac{(1-\alpha)}{2} \ \operatorname{erf}\left(\frac{nx}{\sqrt{2}}\right)+\frac{n(1-\alpha)}{\sqrt{2\pi}} \ x \ e^{-{\frac{n^2x^2}{2}}}
\end{align}
\begin{align}
    \frac{\partial G}{\partial \alpha} = \frac{x}{2}\left(1 - \operatorname{erf}\left(\frac{nx}{\sqrt{2}}\right)\right).
\end{align}
where
\begin{align}
\frac{d}{dx}
    \text{erf}(x) = & \ \frac{2}{\sqrt{\pi}}e^{-x^2}\notag
\end{align}
$\alpha$ and $n$ can be either hyperparameters or trainable parameters.\\ 
Note that the class of neural networks with SAU activation function is dense in $C(K)$, where $K$ is a compact subset of $\mathbb{R}^n$ and $C(K)$ is the space of all continuous functions over $K$.\\
The proof follows from the following propositions (see \cite{pau}) as the proposed function is not a polynomial. 

\textbf{Proposition 1. (Theorem 1.1 in Kidger and Lyons, 2020 \cite{universal}) :-} Let $\rho: \mathbb{R}\rightarrow \mathbb{R}$ be any continuous function. Let $N_n^{\rho}$ represent the class of neural networks with activation function $\rho$, with $n$ neurons in the input layer, one neuron in the output layer, and one hidden layer with an arbitrary number of neurons. Let $K \subseteq \mathbb{R}^n$ be compact. Then $N_n^{\rho}$ is dense in $C(K)$ if and only if $\rho$ is non-polynomial.

\section{Experiments}
We have considered eight popular standard activation functions to compare performance with SAU on different datasets and models on standard deep learning problems like image classification, object detection, semantic segmentation, and machine translation. It is evident from the experimental results in nest sections that SAU outperform in most cases compared to the standard activations. For the rest of our experiments, we have considered $\alpha$ as a trainable parameter and $n$ as a hyper-parameter. We have initialized $\alpha$ at $0.15$ and updated it via backpropagation according to (5). The value of $n$ is considered $20000$. All the experiments are conducted on an NVIDIA V100 GPU with 32GB RAM.\\

\subsection{Image Classification}
\subsubsection{MNIST, Fashion MNIST and The Street View House Numbers (SVHN) Database}In this section, we present results on MNIST \cite{mnist}, Fashion MNIST \cite{fashion}, and SVHN \cite{SVHN} datasets. The MNIST and Fashion MNIST databases have a total of 60k training and 10k testing $28\times 28$ grey-scale images with ten different classes. SVHN consists of $32\times 32$ RGB images with a total of 73257 training images and  26032 testing images with ten different classes. We do not apply any data augmentation method in MNIST and Fashion MNIST datasets. We have applied standard data augmentation methods like rotation, zoom, height shift, shearing on the SVHN dataset. We report results with VGG-16 \cite{vgg} (with batch-normalization) network in Table~\ref{tabic}. 
\begin{table}[H]
\begin{center}
\begin{tabular}{ |c|c|c|c|c| }
 \hline
 Activation Function &  \makecell{MNIST} & \makecell{Fashion MNIST} & \makecell{SVHN}  \\
 \hline
 ReLU  &  $99.05 \pm 0.11$ & $93.13 \pm 0.23$ & $95.09\pm 0.26$\\ 
 \hline
  Swish  & $99.16\pm 0.09$ & $93.34\pm 0.21$ & $95.29\pm 0.20$\\
 \hline
 Leaky ReLU($\alpha$ = 0.01) & $99.02 \pm 0.14$ & $93.17\pm 0.28$ & $95.24\pm 0.23$\\
 \hline
 ELU  & $99.01\pm 0.15$ & $93.12\pm 0.30$ & $95.15\pm 0.28$\\
 \hline
 Softplus & $98.97\pm 0.14$ & $92.98\pm 0.34$ & $94.94\pm 0.30$\\

 \hline
GELU & 99.18$\pm$0.09 & 93.41$\pm$0.29 & 95.11 $\pm$0.24\\
 \hline
 PReLU & 99.01$\pm$0.09 & 93.12$\pm$0.27 & 95.14 $\pm$0.24\\
 \hline
 ReLU6 & 99.20$\pm$0.08 & 93.25$\pm$0.27 & 95.22 $\pm$0.20\\
 \hline
 SAU & \textbf{99.28}$\pm$0.06 & \textbf{93.57}$\pm$0.20 & \textbf{95.41}$\pm$0.18\\
 \hline

 \end{tabular}
 \vspace{0.2cm}
\caption{Comparison between different baseline activations, SAU activation on MNIST, Fashion MNIST, and SVHN datasets on VGG-16 network. We report results for 10-fold mean accuracy (in \%). mean$\pm$std is reported in the table.} 
\label{tabic}
\end{center}
\end{table}

\subsubsection{CIFAR}
The CIFAR \cite{cifar10} is one of the most popular databases for image classification consists of a total of 60k $32\times 32$ RGB images and is divided into 50k training and 10k test images. CIFAR has two different datasets- CIFAR10 and CIFAR100 with a total of 10 and 100 classes, respectively. We report the top-1 accuracy on Table~\ref{tab2} and Table~\ref{tab3} on CIFAR10 dataset CIFAR100 datasets respectively. We consider MobileNet V2 (MN V2) \cite{mobile}, Shufflenet V2 (SF V2) \cite{shufflenet}, and EfficientNet B0 (EN-B0) \cite{efficientnet}. For all the experiments to train a model on these two datasets, we use a batch size of 128, stochastic gradient descent (\cite{sgd1}, \cite{sgd2}) optimizer with 0.9 momentum \& $5e^{-4}$ weight decay, and trained all networks up-to 200 epochs. We begin with 0.01 learning rate and decay the learning rate with cosine annealing \cite{cosan} learning rate scheduler. We have applied Standard data augmentation methods like width shift, height shift, horizontal flip, and rotation on both datasets. It is noticeable from these two tables that replacing ReLU by SAU, there is an increment in top-1 accuracy from 1\% to more than 5\%.

\begin{table}[htbp]
\makebox[\textwidth][c]{
    \begin{tabular}{c|c|c}
\toprule
        Model    &\multicolumn{1}{c|}{ReLU} & \multicolumn{1}{c}{SAU} \\
\midrule
 &  Top-1 accuracy (mean$\pm$ std)     & Top-1 accuracy (mean $\pm$ std)\\
\hline
EffitientNet B0 &  95.01 $\pm$ 0.12 &  \textbf{96.07} $\pm$ 0.08  \\
\midrule 
MobileNet V2 &    94.07 $\pm$ 0.15  &   \textbf{95.25} $\pm$ 0.10\\
\bottomrule
\end{tabular}%
}
\caption{Experimental results on CIFAR10 dataset. Top-1 accuracy(in $\%$) for mean of 8 different runs have been reported. mean$\pm$std is reported in the table.}
\label{tab2}%
\end{table}%


\begin{table}[htbp]
\makebox[\textwidth][c]{
    \begin{tabular}{c|c|c}
\toprule
        Model    &\multicolumn{1}{c|}{ReLU} & \multicolumn{1}{c}{SAU} \\
\midrule
 &  Top-1 accuracy (mean$\pm$ std)     & Top-1 accuracy (mean $\pm$ std)\\
\hline
Shufflenet V2 0.5x  &  62.02 $\pm$ 0.20 &     \textbf{64.39} $\pm$ 0.17\\
Shufflenet V2 1.0x & 64.57 $\pm$ 0.25  & \textbf{68.18} $\pm$ 0.16\\
Shufflenet V2 2.0x &     67.11 $\pm$ 0.23 & \textbf{72.23} $\pm$ 0.16  \\
\bottomrule
\end{tabular}%
}
\caption{Experimental results on CIFAR100 dataset. Top-1 accuracy(in $\%$) for mean of 8 different runs have been reported. mean$\pm$std is reported in the table.}
\label{tab3}
\end{table}
\begin{figure}[H]
    \begin{minipage}[t]{.495\linewidth}
        \centering
    
        \includegraphics[width=\linewidth]{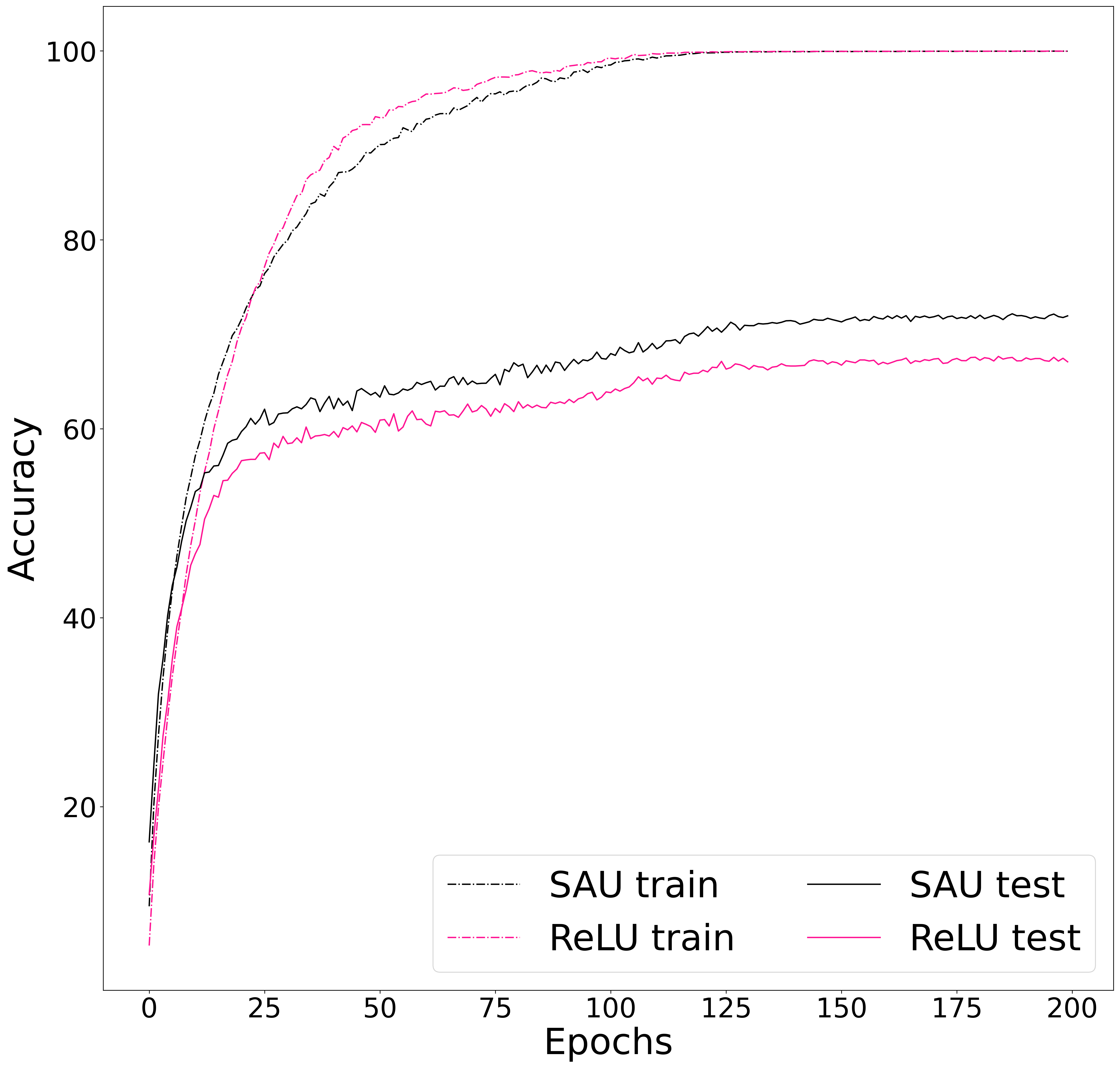}
        \vspace{0.2cm}
        \caption{Top-1 Train and Test accuracy (higher is better) on CIFAR100 dataset with ShuffleNet V2 (2.0x) model for SAU and ReLU.}
        \label{acc2}
    \end{minipage}
    \hfill
    \begin{minipage}[t]{.48\linewidth}
        \centering
        
       \includegraphics[width=\linewidth]{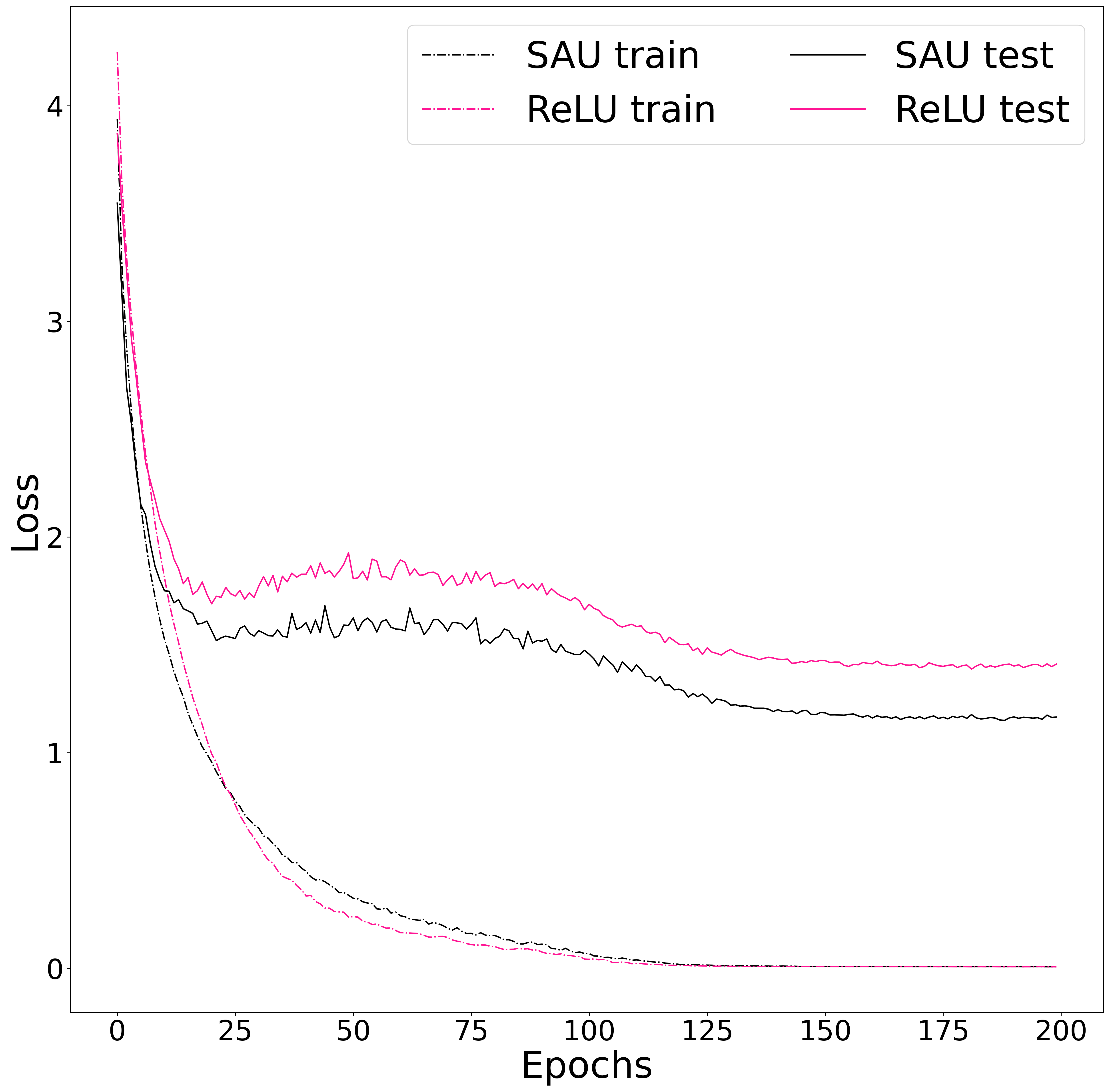}
       \vspace{0.2cm}
        \caption{Top-1 Train and Test loss (lower is better) on CIFAR100 dataset with ShuffleNet V2 (2.0x) model for SAU and ReLU.}
        \label{loss2}
    \end{minipage}  
\end{figure}

\subsubsection{Tiny Imagenet}
 In this section, we present result on Tiny ImageNet dataset, a similar kind of image classification database like the ImageNet Large Scale Visual Recognition Challenge(ILSVRC). Tiny Imagenet contains $64 \times 64$ RGB images with total 100,000 training images, 10,000 validation images, and 10,000 test images and have total 200 image classes. We report a mean of 5 different runs for Top-1 accuracy in table~\ref{tab22222} on WideResNet 28-10 (WRN 28-10) \cite{wrn} model. We consider a batch size of 32, He Normal initializer \cite{prelu}, 0.2 dropout rate \cite{dropout}, adam optimizer \cite{adam}, initial learning rate(lr rate) 0.01, and lr rate is reduced by a factor of 10 after every 60 epochs up-to 300 epochs. We have applied standard data augmentation methods like rotation, width shift, height shift, shearing, zoom, horizontal flip, fill mode. It is evident from the table that the proposed function performs better than the baseline functions, and top-1 accuracy is stable (mean$\pm$std) and got a good improvement for SAU over ReLU.

\begin{table}[H]
\begin{center}
\begin{tabular}{ |c|c|c| }
 \hline
 Activation Function &  \makecell{Wide ResNet \\ 28-10 Model}  \\
 \hline
 ReLU  &  61.61$\pm$0.47  \\ 
 \hline

 Swish  &  62.44$\pm$0.49\\
  \hline
 Leaky ReLU($\alpha$ = 0.01) &  61.47$\pm$0.44 \\
 \hline
 ELU  &  61.99$\pm$0.57\\
 \hline
 Softplus & 60.42$\pm$0.61\\

 \hline
 GELU & 62.64$\pm$0.62\\
 \hline

 PReLU & 61.25$\pm$0.51\\
 \hline
 ReLU6 & 61.72$\pm$0.56\\
  \hline
  
 SAU &  \textbf{63.20}$\pm$0.51\\ 

  \hline
 \end{tabular}
 \vspace{0.4cm}
\caption{Experimental results on Tiny ImageNet dataset. Mean of 5 different runs for top-1 accuracy(in $\%$) have been reported. mean$\pm$std is reported in the table.} 
\label{tab22222}
\end{center}
\end{table}
\subsection{Object Detection}
A standard problem in computer vision is object detection, in which the network model try to locate and identify each object present in the image. Object detection is widely used in face detection, autonomous vehicle etc. In this section, we present our results on challenging Pascal VOC dataset \cite{pascal} on Single Shot MultiBox Detector(SSD) 300 \cite{ssd} with VGG-16(with batch-normalization) \cite{vgg} as the backbone network. No pre-trained weight is considered for our experiments in the network. The network has been trained with a batch size of 8, 0.001 learning rate, SGD optimizer \cite{sgd1, sgd2} with 0.9 momentum, 5$e^{-4}$ weight decay for 120000 iterations. We report the mean average precision (mAP) in Table~\ref{tabod} for a mean of 5 different runs.

\begin{table}[H]
\begin{center}
\begin{tabular}{ |c|c|c| }
 \hline
 Activation Function &  \makecell{mAP}  \\
  
 \hline
 ReLU  &  77.2$\pm$0.14  \\ 
 \hline
 Swish  &  77.3$\pm$0.11\\
 \hline
 Leaky ReLU($\alpha$ = 0.01) &  77.2$\pm$0.19 \\
 \hline
 ELU  &  75.1$\pm$0.22\\
 \hline
 Softplus & 74.2$\pm$0.25\\
 \hline

 GELU & 77.3$\pm$0.12\\
 \hline
PReLU & 77.2$\pm$0.20\\
\hline
ReLU6 & 77.1$\pm$0.15\\
\hline
SAU &  \textbf{77.7}$\pm$0.10 \\ 

\hline
 \end{tabular}
 \vspace{0.2cm}
\caption{Object Detection results on SSD 300 model in Pascal-VOC dataset. mean$\pm$std is reported in the table.} 
\label{tabod}
\end{center}
\end{table}

\subsection{Semantic Segmentation}
Semantic segmentation is a computer vision problem that narrates the procedure of associating each pixel of an image with a class label. We present our experimental results in this section on the popular Cityscapes dataset \cite{city}. The U-net model \cite{unet} is considered as the segmentation framework and is trained up-to 250 epochs, with adam optimizer \cite{adam}, learning rate 5$e^{-3}$, batch size 32 and Xavier Uniform initializer \cite{xavier}. We report the mean of 5 different runs for Pixel Accuracy and mean Intersection-Over-Union (mIOU) on test data on table~\ref{tabsg}.
\begin{table}[H]
\begin{center}
\begin{tabular}{ |c|c|c|c| }
 \hline
 Activation Function & \makecell{
Pixel\\ Accuracy} & mIOU  \\
 \hline
 ReLU  & 79.60$\pm$0.45 & 69.32$\pm$0.30 \\ 
 \hline
 Swish  & 79.71$\pm$0.49 & 69.68$\pm$0.31\\
 \hline
 Leaky ReLU($\alpha$ = 0.01) & 79.41$\pm$0.42 & 69.48$\pm$0.39\\
 \hline
 ELU  & 79.27$\pm$0.54 & 68.12$\pm$0.41\\
 \hline
 Softplus & 78.69$\pm$0.49 & 68.12$\pm$0.55\\
 \hline

 GELU & 79.60$\pm$0.39 & 69.51$\pm$0.39\\
 \hline
 PReLU & 78.99$\pm$0.42 & 68.82$\pm$0.41\\
 \hline
 ReLU6 & 79.59$\pm$0.41 & 69.66$\pm$0.41\\
 \hline
SAU & \textbf{81.11}$\pm$0.40 & \textbf{71.02}$\pm$0.32 \\ 

 \hline
 \end{tabular}
 \vspace{0.3cm}
\caption{semantic segmentation results on U-NET model in CityScapes dataset. mean$\pm$std is reported in the table.} 
\label{tabsg}
\end{center}
\end{table}

\subsection{Machine Translation}
Machine Translation is a deep learning technique in which a model translate text or speech from one language to another language. In this section, we report results on WMT 2014 English$\rightarrow$German dataset. The database contains 4.5 million training sentences. Network performance is evaluated on the newstest2014 dataset using the BLEU score metric. An Attention-based 8-head transformer network \cite{attn} is used with Adam optimizer \cite{adam}, 0.1 dropout rate \cite{dropout}, and trained up to 100000 steps. We try to keep other hyper-parameters similar as mentioned in the original paper \cite{attn}.  We report mean of 5 runs on Table~\ref{tabmt} on the test dataset(newstest2014).
\begin{table}[H]
\begin{center}
\begin{tabular}{ |c|c|c| }
 \hline
 Activation Function & \makecell{
BLEU Score on\\ the newstest2014 dataset }  \\
 
 \hline
 ReLU  &  26.2$\pm$0.15  \\ 
 \hline
 Swish  &  26.4$\pm$0.10\\
 \hline
 Leaky ReLU($\alpha$ = 0.01) &  26.3$\pm$0.17 \\
 \hline
 ELU  &  25.1$\pm$0.15\\
 \hline
 Softplus & 23.6$\pm$0.16\\
 \hline
 GELU & 26.4$\pm$0.19\\
 \hline
  PReLU & 26.2$\pm$0.21\\
 \hline
 ReLU6 & 26.1$\pm$0.14\\
 \hline
 SAU &  \textbf{26.7}$\pm$0.12 \\ 
 
 \hline
 \end{tabular}
 \vspace{0.2cm}
\caption{Machine translation results on transformer model in WMT-2014 dataset. mean$\pm$std is reported in the table.} 
\label{tabmt}
\end{center}
\end{table}

\section{Baseline Table}

In this section, we present a table for SAU and the other baseline functions, which shows that  SAU beat or perform equally well compared to baseline activation functions in most cases. We present a detailed comparison based on all the experiments in earlier sections with SAU and the baseline activation functions in Table~\ref{tab49}.

\begin{table}[H]
\newenvironment{amazingtabular}{\begin{tabular}{*{50}{l}}}{\end{tabular}}
\centering
\begin{amazingtabular}
\midrule
Baselines & ReLU & \makecell{Leaky\\ ReLU} & ELU & Softplus & Swish  & PReLU & ReLU6 & GELU\\
\midrule
SAU $>$ \text{Baseline} & \hspace{0.3cm}12 & \hspace{0.3cm}12 & \hspace{0.3cm}12 & \hspace{0.3cm}12 & \hspace{0.3cm}12 & \hspace{0.3cm}12 & \hspace{0.3cm}12 & \hspace{0.3cm}12 \\
SAU $=$ \text{Baseline} & \hspace{0.3cm}0 & \hspace{0.3cm}0 & \hspace{0.3cm}0 & \hspace{0.3cm}0 & \hspace{0.3cm}0 & \hspace{0.3cm}0  & \hspace{0.3cm}0  & \hspace{0.3cm}0 \\
SAU $<$ \text{Baseline} & \hspace{0.3cm}0 & \hspace{0.3cm}0 & \hspace{0.3cm}0 & \hspace{0.3cm}0 & \hspace{0.3cm}0 & \hspace{0.3cm}0 & \hspace{0.3cm}0 & \hspace{0.3cm}0 \\

\bottomrule
\end{amazingtabular}
\vspace{0.2cm}
  \caption{Baseline table for SAU. These numbers represent the total number of models in which SAU underperforms, equal or outperforms compared to the baseline activation functions}
  \label{tab49}
\end{table}

\section{Conclusion}
In this paper, we propose a new novel smooth activation function using approximate identity, and we call it smooth activation unit (SAU). The proposed function can approximate ReLU or its different variants (like Leaky ReLU etc.). We show that on a wide range of experiments on different deep learning problems, the proposed functions outperform the known activations like ReLU or Leaky ReLU in most cases. 

\bibliographystyle{unsrt} 
\bibliography{references.bib}
\end{document}